\documentclass{article}
\usepackage[utf8]{inputenc}
\usepackage{main}
\usepackage{microtype}
\usepackage{graphicx}
\usepackage{subfig}
\usepackage{times}
\usepackage{latexsym}
\usepackage{amsmath}
\usepackage{float}
\usepackage{footnote}
\usepackage{enumitem}
\usepackage{bm}
\usepackage{arydshln}
\usepackage{booktabs}
\usepackage{tabularx}
\usepackage{multicol}
\usepackage{multirow}
\usepackage{color}
\usepackage{xcolor}     
\usepackage{colortbl}
\usepackage{bbding}
\usepackage{makecell}
\usepackage{mathtools}
\usepackage{imakeidx}
\usepackage{longtable}
\usepackage{wrapfig}
\makeindex
\usepackage{arydshln}
\usepackage{lipsum}
\usepackage{natbib}
\usepackage[toc]{multitoc}
\usepackage[edges]{forest}
\usepackage[normalem]{ulem}
\definecolor{mydarkblue}{rgb}{0,0.08,0.45}
\usepackage[colorlinks=true,linkcolor=mydarkblue,citecolor=mydarkblue,filecolor=mydarkblue,urlcolor=mydarkblue]{hyperref}
\usepackage{caption}
\usepackage{CJKutf8}
\usepackage{awesomebox} 
\usepackage{bbding}
\usepackage{epigraph}
\usepackage{csquotes}
\usepackage[most]{tcolorbox}
\usepackage{amssymb}
\usepackage{array}
\usepackage{threeparttable}
\usepackage{ragged2e}

\usepackage{pifont}   

\usepackage[tikz]{bclogo}
\usepackage[framemethod=tikz]{mdframed}
\definecolor{bgblue}{RGB}{245,243,253}
\definecolor{ttblue}{RGB}{91,194,224}

\mdfdefinestyle{mystyle}{%
  rightline=true,
  innerleftmargin=10,
  innerrightmargin=10,
  outerlinewidth=3pt,
  topline=false,
  rightline=true,
  bottomline=false,
  skipabove=\topsep,
  skipbelow=\topsep
}

\newtcolorbox{myboxi}[1][]{
  breakable,
  title=#1,
  colback=red!5,
  colbacktitle=red!5,
  coltitle=black,
  fonttitle=\bfseries,
  bottomrule=0pt,
  toprule=0pt,
  leftrule=2pt,
  rightrule=2pt,
  titlerule=0pt,
  arc=0pt,
  outer arc=0pt,
  colframe=red,
}

\newtcolorbox{myboxnote}[1][]{
  breakable,
  title=#1,
  colback=orange!0,
  colbacktitle=orange!0,
  coltitle=black,
  fonttitle=\bfseries,
  bottomrule=0pt,
  toprule=0pt,
  leftrule=2pt,
  rightrule=2pt,
  titlerule=0pt,
  arc=0pt,
  outer arc=0pt,
  colframe=orange,
}

\newtcolorbox{myboxii}[1][]{
  breakable,
  freelance,
  title=#1,
  colback=white,
  colbacktitle=white,
  coltitle=black,
  fonttitle=\bfseries,
  bottomrule=0pt,
  boxrule=0pt,
  colframe=white,
  overlay unbroken and first={
  \draw[red!75!black,line width=3pt]
    ([xshift=5pt]frame.north west) -- 
    (frame.north west) -- 
    (frame.south west);
  \draw[red!75!black,line width=3pt]
    ([xshift=-5pt]frame.north east) -- 
    (frame.north east) -- 
    (frame.south east);
  },
  overlay unbroken app={
  \draw[red!75!black,line width=3pt,line cap=rect]
    (frame.south west) -- 
    ([xshift=5pt]frame.south west);
  \draw[red!75!black,line width=3pt,line cap=rect]
    (frame.south east) -- 
    ([xshift=-5pt]frame.south east);
  },
  overlay middle and last={
  \draw[red!75!black,line width=3pt]
    (frame.north west) -- 
    (frame.south west);
  \draw[red!75!black,line width=3pt]
    (frame.north east) -- 
    (frame.south east);
  },
  overlay last app={
  \draw[red!75!black,line width=3pt,line cap=rect]
    (frame.south west) --
    ([xshift=5pt]frame.south west);
  \draw[red!75!black,line width=3pt,line cap=rect]
    (frame.south east) --
    ([xshift=-5pt]frame.south east);
  },
}

\usepackage{fancyhdr} 
\usepackage{blindtext} 

\DeclareCaptionFont{black}{\color{black}}

\definecolor{myblue}{rgb}{0.9, 0.1, 0.94}
\definecolor{mygreen}{rgb}{0.64, 0.56, 0.88}
\definecolor{myyellow}{rgb}{0.68, 0.6, 0.1}
\definecolor{fancygreen}{rgb}{0.33, 0.68, 0.20}
\definecolor{salmon}{rgb}{0.94, 0.52, 0.49}
\definecolor{tablegreen}{rgb}{0.82, 0.94, 0.75}
\definecolor{tableblue}{rgb}{0.81, 0.90, 0.94}
\definecolor{tablered}{rgb}{0.97, 0.85, 0.85}
\definecolor{tableorange}{rgb}{0.96, 0.85, 0.81}



\newenvironment{itemize*}%
 {\leftmargini=10pt\begin{itemize}%
  \setlength{\itemsep}{0pt}%
  \setlength{\parskip}{0pt}%
  }%
 {\end{itemize}}
\newenvironment{enumerate*}%
 {\begin{enumerate}%
  \setlength{\itemsep}{0pt}%
  \setlength{\parskip}{0pt}}%
 {\end{enumerate}}

\usepackage{xcolor}
\usepackage{listings}

\newcommand\JSONnumbervaluestyle{\color{blue}}
\newcommand\JSONstringvaluestyle{\color{red}}

\newif\ifcolonfoundonthisline

\makeatletter

\lstdefinestyle{json}
{
  showstringspaces    = false,
  keywords            = {false,true},
  alsoletter          = 0123456789.,
  morestring          = [s]{"}{"},
  stringstyle         = \ifcolonfoundonthisline\JSONstringvaluestyle\fi,
  MoreSelectCharTable =%
    \lst@DefSaveDef{`:}\colon@json{\processColon@json},
  basicstyle          = \ttfamily,
  keywordstyle        = \ttfamily\bfseries,
}

\newcommand\processColon@json{%
  \colon@json%
  \ifnum\lst@mode=\lst@Pmode%
    \global\colonfoundonthislinetrue%
  \fi
}

\lst@AddToHook{Output}{%
  \ifcolonfoundonthisline%
    \ifnum\lst@mode=\lst@Pmode%
      \def\lst@thestyle{\JSONnumbervaluestyle}%
    \fi
  \fi
  \lsthk@DetectKeywords%
}

\lst@AddToHook{EOL}%
  {\global\colonfoundonthislinefalse}

\makeatother

\usepackage{etoolbox}
\usepackage{natbib}
\usepackage{url}
\newcounter{bibcount}
\makeatletter
\patchcmd{\@lbibitem}{\item[}{\item[\hfil\stepcounter{bibcount}{[\thebibcount]}}{}{}
\renewcommand\NAT@bibsetup%
  [1]{\setlength{\leftmargin}{\bibhang}\setlength{\itemindent}{-\parindent}%
      \setlength{\itemsep}{\bibsep}\setlength{\parsep}{\z@}}
\makeatother

\newcommand*\samethanks[1][\value{footnote}]{\footnotemark[#1]}

\author{Ethan Chern\textsuperscript{\rm{1,4}}\thanks{Equal Contribution.}~\thanks{Partial work done at Bytedance Seed.}\space\space\space
Zhulin Hu\textsuperscript{\rm{1,4}}\samethanks[1]\space\space \space
Steffi Chern\textsuperscript{\rm{4}}\samethanks[1]\space\space \space
Siqi Kou\textsuperscript{\rm{1}}\space\space\space
Jiadi Su\textsuperscript{\rm{3,4}} \space\space\space
Yan Ma\textsuperscript{\rm{3,4}}\\
\textbf{Zhijie Deng}\textsuperscript{\rm{1}}\space\space \space\space \space\space
\textbf{Pengfei Liu}\textsuperscript{\rm{1,2,4}}\thanks{Corresponding author.}\\
    \textsuperscript{\rm 1}Shanghai Jiao Tong University\space\space
    \textsuperscript{\rm 2}SII \space\space \\
    \textsuperscript{\rm 3}Fudan University \space\space
    \textsuperscript{\rm 4}Generative AI Research Lab (GAIR)
}

\begin{document}
\title{Thinking with Generated Images} 
 
\maketitle
\thispagestyle{fancy}
\fancyhead{}
\lhead{\includegraphics[height=0.67cm]{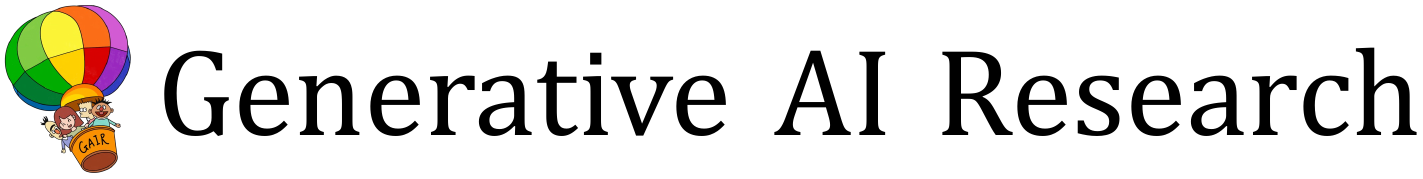}} \renewcommand{\headrulewidth}{0.0pt}
\setlength{\headheight}{0.42cm}

\begin{abstract}
  We present \textit{Thinking with Generated Images}, a novel paradigm that fundamentally transforms how large multimodal models (LMMs) engage with visual reasoning by enabling them to \textit{natively} think across text and vision modalities through \textit{spontaneous} generation of intermediate visual thinking steps. Current visual reasoning with LMMs is constrained to either processing fixed user-provided images or reasoning solely through text-based chain-of-thought (CoT). \textit{Thinking with Generated Images} unlocks a new dimension of cognitive capability where models can actively construct intermediate visual thoughts, critique their own visual hypotheses, and refine them as integral components of their reasoning process. To implement \textit{Thinking with Generated Images}, we introduce the \textit{native long-multimodal thought process}, which enables unified LMMs to seamlessly generate intermediate visual thoughts, establish visual subgoals, and iteratively critique their visual hypotheses within a single, coherent reasoning process. This approach naturally performs test-time scaling across modalities.
  We demonstrate the effectiveness of our approach through two complementary mechanisms: (1) vision generation with intermediate visual subgoals, where models decompose complex visual tasks into manageable components that are generated and integrated progressively, and (2) vision generation with self-critique, where models generate an initial visual hypothesis, analyze its shortcomings through textual reasoning, and produce refined outputs based on their own critiques. Our experiments on vision generation benchmarks show substantial improvements over baseline approaches, with our models achieving up to 50\% (from 38\% to 57\%) relative improvement in handling complex multi-object scenarios. Beyond these immediate performance gains, \textit{Thinking with Generated Images} opens transformative possibilities for AI systems across diverse real-world applications. From biochemists exploring novel protein structures, and architects iterating on spatial designs, to forensic analysts reconstructing crime scenes, and basketball players envisioning strategic plays, our approach enables AI models to engage in the kind of visual imagination and iterative refinement that characterizes human creative, analytical, and strategic thinking. This work establishes a foundation for future research in multimodal cognition and complex visual reasoning tasks. We release our open-source suite at \url{https://github.com/GAIR-NLP/thinking-with-generated-images}.

\end{abstract}

\begin{figure}[h]
    \centering
    \includegraphics[width=0.99\textwidth, height=5.2cm]{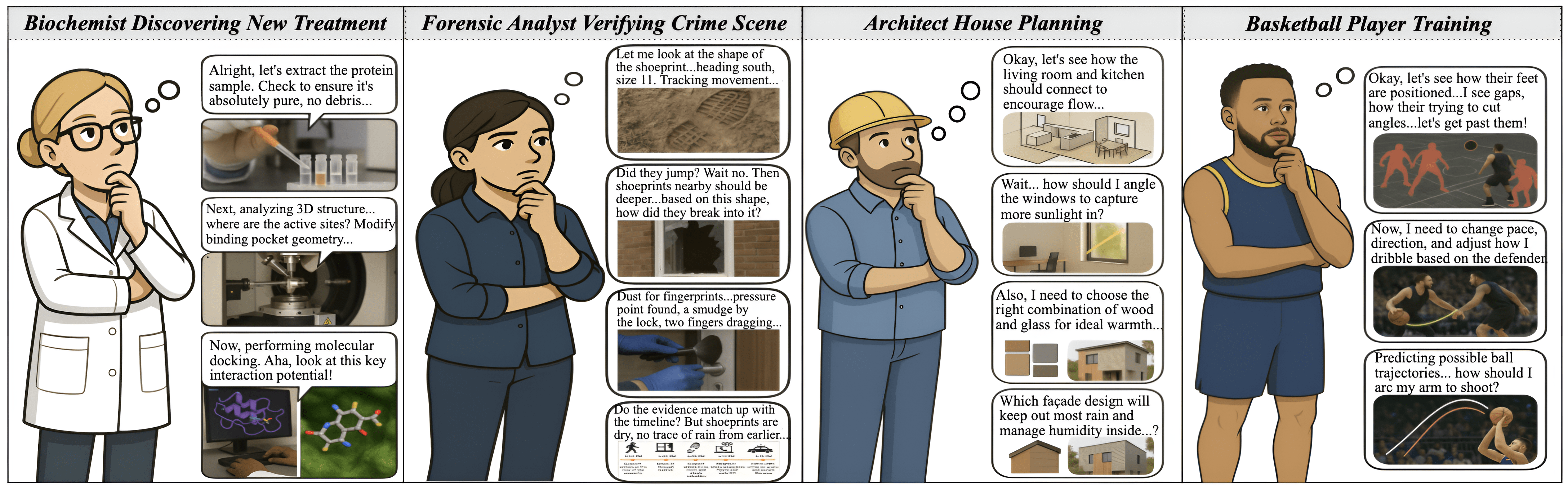}
    \caption{Real-world tasks that require \textit{Thinking with Generated Images}. These tasks often require visual foresight and imagination, which text-based thought alone cannot fully accomplish.}
    \label{fig:framework}
\end{figure}

\newpage

\pagestyle{fancy}
\lhead{\rightmark}
\renewcommand{\headrulewidth}{0.7pt}
\setlength{\headsep}{5mm}

\section{Introduction}
\begin{flushleft}
\leftskip=1cm\emph{``Thorough thought leads to victory; insufficient thought leads to defeat.''} \\
\vspace{.3em}
\leftskip=12.5cm---\emph{Sun Tzu}
\end{flushleft}

Human goal‑oriented cognition \citep{simon1971human, newell2014reasoning, xia2025generativeaiactii} can be characterized as a goal‑directed \textit{search} through a \textit{problem space}\footnote{A problem space is the basic unit of goal‑oriented symbolic activity \citep{newell2014reasoning}. It consists of states (the possible configurations of the system), operators (actions that transform one state into another), an initial state, one or more goal states, and optional path constraints that restrict legal moves. Take the \textit{Tower of Hanoi} as an example: a state is any arrangement of the disks on three pegs; the operators are “move the top disk from one peg to another” and “recognizing the arrangement of the disks”; the initial state has all disks stacked on peg1; the goal state has them stacked on peg3; and the path constraint forbids placing a larger disk atop a smaller one.}—a landscape of intermediate states connected by operators (action or thinking steps). Large language models (LLMs) exhibit this same pattern: when prompted to write a chain‑of‑thought (CoT) \citep{wei2022chain}, they traverse intermediate states, with performance improving with additional inference compute—a.k.a. test‑time scaling \citep{brown2024large,snell2024scaling}. Yet, the text-only CoT process of LLMs captures only a partial view of cognitive search.

Crucially, human cognition is natively \textit{multimodal}. Biochemists explore protein structures to discover new treatment approaches; forensic analysts verify crime scene reconstructions to establish evidence connections; architects revise spaces and light patterns to optimize building designs. Visual thinking creates unique combinations and novel connections between concepts \citep{hao2024training, barrault2024large}. This helps us glimpse new possibilities in our thinking—ideas and connections we wouldn't have discovered through text reasoning alone.

An AI system that thinks or reasons only in the text modality is therefore constrained:\footnote{We argue that vision-language models that are only capable of generating text are still reasoning only in the text modality—they cannot construct intermediate visual thinking.} it cannot explore alternative therapeutic pathways as freely, verify criminal evidence at a glance, or revise design configurations in place\footnote{Although visual content can sometimes be encoded into pure text descriptions, this process is usually inevitably lossy, except for certain tasks like maze problem solving where states and steps can be fully encoded with code or text.}—capacities that underlie much of human creativity and analysis. Extending the thoughts of AI systems to multiple modalities unlocks new forms of interactivity: a model can generate not only intermediate text but also visual thoughts, decompose tasks into both text and visual subgoals, critique these intermediate steps, and iterate—ideally, all within a single, coherent thought process.

While standard large multimodal models (LMMs) or vision-language models (VLMs) can process visual information, these models are merely \textit{seeing} images—processing them once during the forward pass rather than \textit{thinking} with images more deeply. To address this limitation, previous works have explored leveraging multi-step or tool-augmented approaches \citep{zhang2023multimodal, gupta2023visual, suris2023vipergpt, hu2024visual, openai2025thinking, su2025openthinkimglearningthinkimages} to enable models to revisit or conduct simple operations on user-input images as intermediate steps toward solutions. However, these works on \textit{Thinking with Images} only enable models to perform limited operations on images, constraining the thinking space of the model.

To establish a generalized, scalable model that can naturally think and reason across modalities while avoiding the limitations mentioned above, we propose the idea of ``\textit{Thinking with Generated Images}'': An AI model (system) that \textit{spontaneously} thinks and reasons across (e.g., text and vision) modalities. Rather than entirely relying on user‑supplied images, which would constrain its thinking, the model proactively produces its own visual steps or subgoals toward solving problems when needed.

Recent works have attempted to take the first steps toward \textit{Thinking with Generated Images}, using either agentic \citep{guo2025can, fang2025got, ni2024generate} or single-model approaches \citep{li2025imagine}. 
We argue that improving AI system performance via agentic approaches is parallel to single-model approaches, but integrating various capabilities into a single, end-to-end generalist model represents the continued evolution toward superintelligence \citep{bubeck2023sparks}. More recent work \citep{li2025imagine}
attempted to leverage unified LMMs (models that can generate multimodal content simultaneously) to solve spatial reasoning tasks such as maze navigation. While taking a first step, their approach is limited to maze problem-solving, which can be adequately solvable without visual intermediate steps \citep{dao2025alphamazeenhancinglargelanguage}. We argue that \textit{Thinking with Generated images} is best applied to complex tasks (as shown in Fig.~\ref{fig:framework}), where their intermediate steps \textit{cannot} be adequately represented in textual form alone. Thus, to demonstrate the potential of \textit{Thinking with Generated Images}, in this paper, we focus on vision (image) generation tasks.

To effectively realize \textit{Thinking with Generated Images} with an end-to-end, single-model approach and demonstrate its strong potential on vision generation tasks, we introduce the \textbf{native long-multimodal thought process} on unified autoregressive LMMs \citep{team2024chameleon, chern2024anole}. The native long-multimodal thought process comprises interleaved multimodal tokens—words or subwords for text, patches for vision, frames for audio, and
other domain-specific representations\footnote{We focus on text and vision modalities in this paper.}. The native long-multimodal thought process not only naturally enable models to spontaneously generate images in the thinking process, but also natively perform test-time scaling for better model capabilities.

In summary, our contributions are as follows:
\begin{itemize}
    \item We propose the concept of \textit{Thinking with Generated Images}, demonstrating how a single AI model can learn to spontaneously think and reason across text and vision modalities.
    \item We explore test-time scaling for native unified LMMs, shedding light on future research directions.
    \item We introduce the \textit{native long-multimodal thought process} to instantiate \textit{Thinking with Generated Images}, enabling models to think and reason across modalities, and naturally performs test-time scaling. We demonstrate that the native long-multimodal thought process is beneficial for vision generation and can potentially be applied to more diverse and complex tasks with the emergence of stronger base models.
\end{itemize}

\section{Thinking with Generated Images}

\begin{figure}[H]
    \centering
  \includegraphics[width=0.73\textwidth, height=6cm]{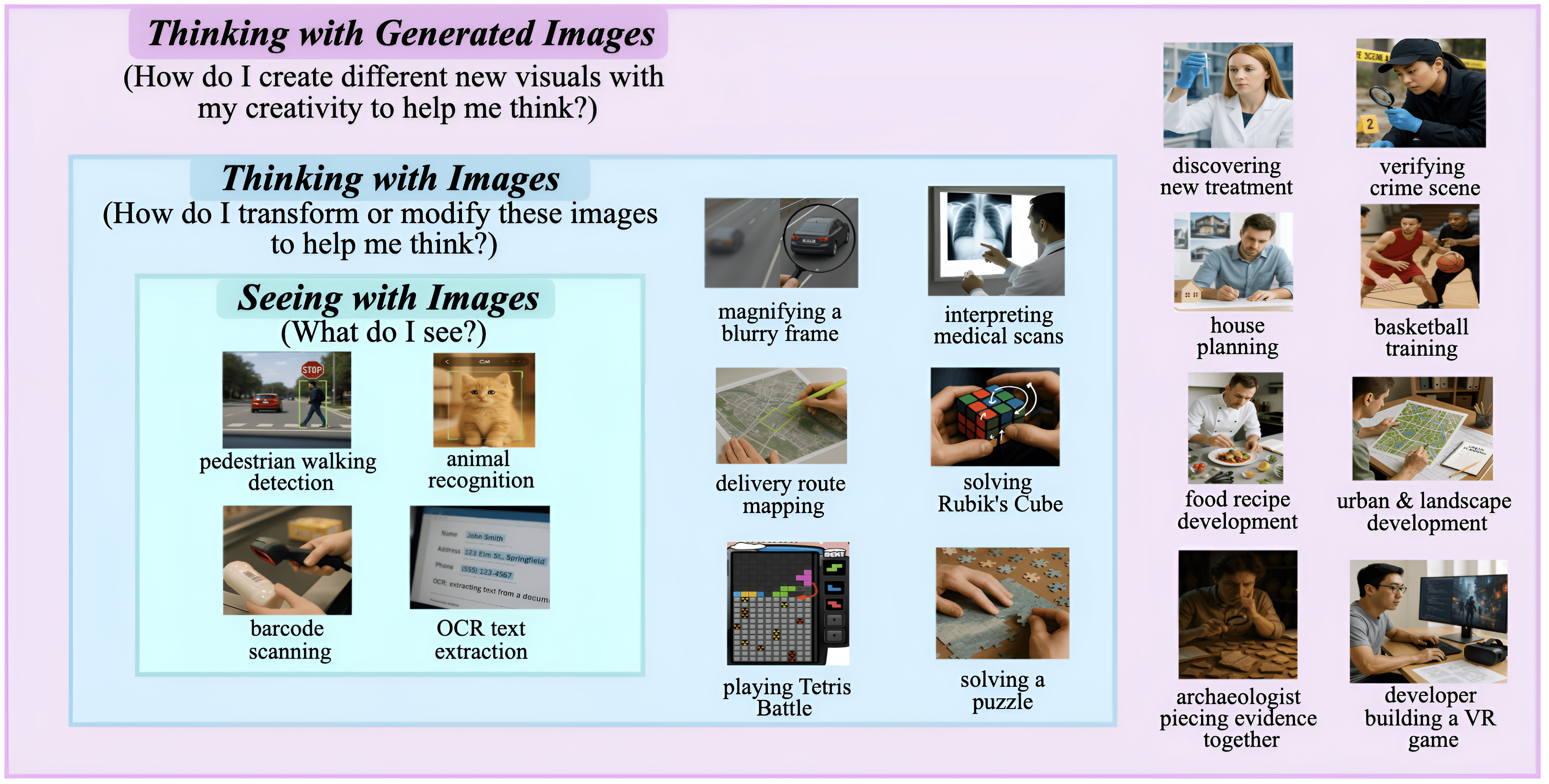}
  \caption{Examples of “seeing with images” vs.\ “thinking with images” vs.\ “thinking with generated images.”}
  \label{fig:see_vs_think}
  \vspace{-7pt}                      
\end{figure}

\subsection{``Seeing'' vs.\ ``Thinking'' with Images}

Standard LMMs or VLMs only \emph{see} an image once.  
During a single forward pass, the image is converted into visual tokens, and the model then autoregressively generates text tokens conditioned on those visual tokens. The thinking process of AI models happens \textit{entirely} in the text modality; the image acts as a fixed prior or conditioning context for a \emph{text‑only} chain‑of‑thought.

To integrate images into the thought process, recent work leverages established multi-step pipelines \citep{zhang2023multimodal}, multi-agent systems \citep{yang2023mm}, or external tools such as code executors, OCR, or image manipulators \citep{gupta2023visual,suris2023vipergpt,hu2024visual,openai2025thinking}.  
These approaches let the model revisit the (user-given) images multiple times or generate transformed versions (e.g., cropped, rotated, zoomed) as intermediate steps toward a solution.

\emph{Thinking} with images—rather than merely \textit{seeing} them once—enables models to better tackle multi‑step visual reasoning tasks such as spatial puzzles, complex visual question answering, and chart or diagram interpretation.

\subsection{Thinking with Images v.s. Thinking with ``Generated'' Images}

Thinking with images alone is not sufficient—models are still constrained when limited to reasoning with fixed user input images or slightly ``transformed'' versions of these images. The reasoning steps for tasks like design planning and physical-world reasoning would be lossy if they entirely relied on text and user-input image transformations.

To attempt the first steps toward thinking with generated images, previous works have primarily relied on multi-component agentic approaches \citep{guo2025can, fang2025got, ni2024generate} that require multiple model passes to achieve pipelined planning, (re)captioning, generation, critique, and reward components. Though demonstrating great potential, the complexity of these systems makes further generalizability and scaling to diverse scenarios complicated. Additionally, established test-time scaling and post-training scaling methods like long-form CoT and reinforcement learning must be engineered separately for each module, a burden that adds unwarranted complexity. These limitations ultimately place a ceiling on the performance potential of such fragmented systems. More recent works \citep{li2025imagine} have attempted single-model, end-to-end approaches. However, their approach limited its tasks to spatial reasoning such as maze problem-solving, where the maze itself and the intermediate states and steps are essentially discrete and can be entirely encoded with pure text descriptions \citep{dao2025alphamazeenhancinglargelanguage}. Such tasks are thus solvable without visual intermediate steps and don't fully unlock the potential of \textit{Thinking with Generated Images}.

\begin{figure}[H]
  \centering
  \includegraphics[width=0.73\textwidth, height=6.5cm]{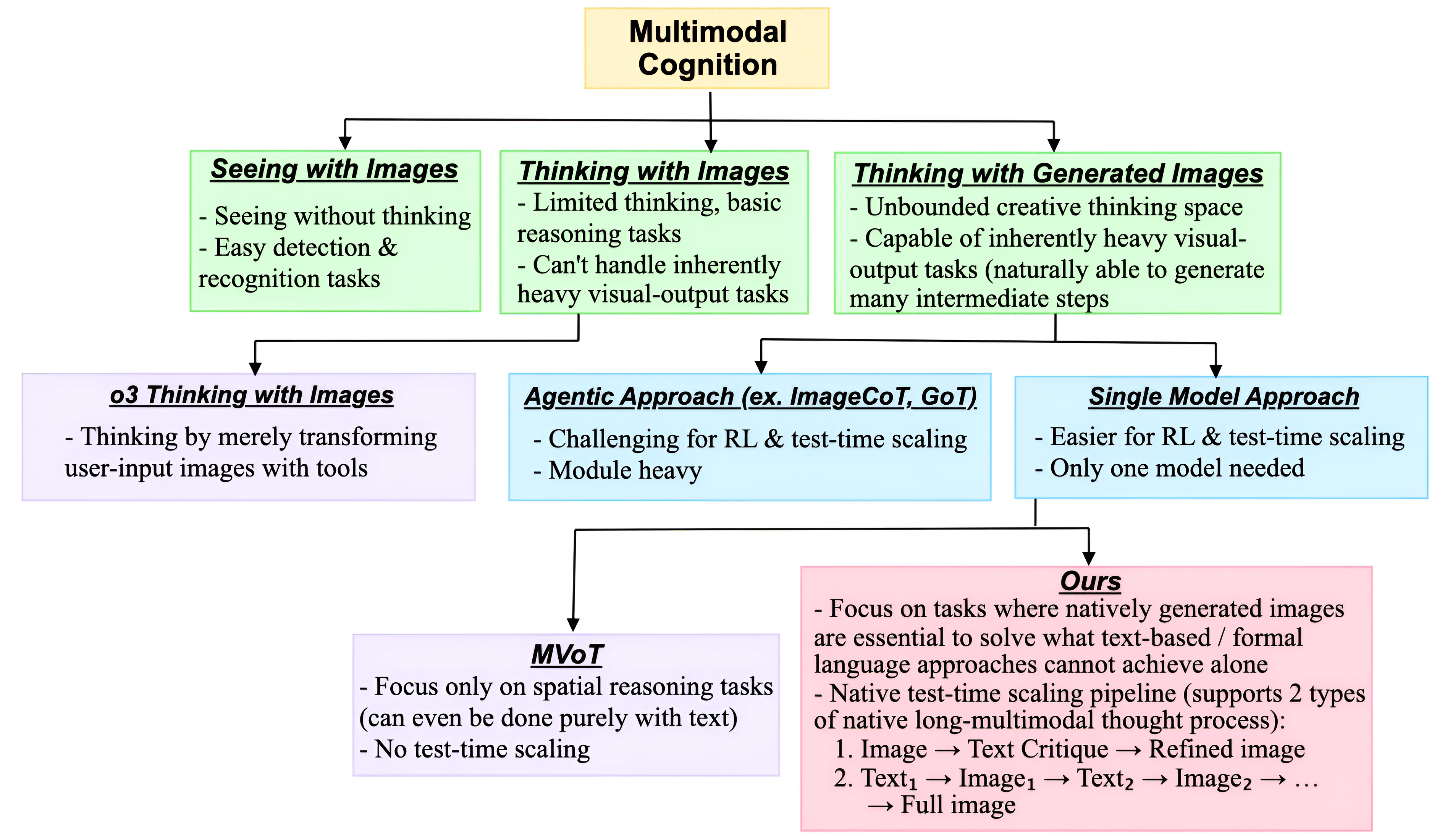}
  \caption{Comparison between related works on multimodal cognition.}
  \label{fig:flow_chart}
  \vspace{-10pt}                
\end{figure}

To fully unlock the potential of \textit{Thinking with Generated Images}, we focus on tasks that fundamentally require intermediate thoughts expressed in visual space. In this paper, we concentrate specifically on vision (image) generation, while demonstrating how our approach (the native long-multimodal thought process) enables models to inherently and deeply integrate vision into their reasoning processes. Unlike approaches that rely on fixed, user-inputted images, \textit{Thinking with Generated Images} allows models to dynamically create and manipulate visual representations throughout their reasoning process. This fundamental shift unlocks new forms of interactivity and enables capabilities such as 3D modeling, visual foresight, design, and intuitive physics—domains where visual intermediate steps are not merely helpful but essential for effective problem-solving. A summary of the differences between our work and previous related works can be found in Fig.~\ref{fig:flow_chart}.

\subsection{Thinking with Generated Images with the Native Long-Multimodal Thought Process}

We implement \textit{Thinking with Generated Images} through the 
\textit{native long-multimodal thought process} that leverages unified autoregressive LLMs via effective supervised fine-tuning on carefully curated high-quality samples. The native long-multimodal thought process offers multi-faceted benefits that transcend previous approaches: (1) enables models to “think’’ across modalities by generating modality-specific tokens with “natively'' embedded multimodal generation capabilities via a single forward pass; (2) performs diverse multimodal tasks natively via the generative paradigm; (3) provides natural test-time scaling across modalities via the generated long-thought; and (4) facilitates future integration with post-training scaling techniques like reinforcement learning.

\subsubsection{The Native Long-Multimodal Thought Process}
The native long-multimodal thought process is a sequence of interleaved multimodal tokens. Formally, if we have $m$ modalities with vocabulary spaces of $\mathcal{V}_1, \mathcal{V}_2, \ldots, \mathcal{V}_m$, we define the unified vocabulary as $\mathcal{V} = \mathcal{V}_1 \cup \mathcal{V}_2 \cup \ldots \cup \mathcal{V}_m$. Importantly, these vocabulary spaces need not be finite sets of discrete tokens (i.e., not constrained to autoregressive next-token-prediction LMMs); they could also be infinite sets of continuous tokens. A sequence of multimodal tokens is represented as $\mathbf{x} = (x_1, x_2, \ldots, x_n)$ where $\forall i, x_i \in \mathcal{V}$. 

Several length or token constraints may be introduced to ensure reasonable interaction within the multimodal thought process. For example, vision tokens might be required to appear in fixed-size blocks (e.g., exactly 1024 consecutive tokens for each image representation), and special delimiter tokens may be introduced to explicitly mark the boundaries and transitions between different modality spaces within the unified vocabulary $\mathcal{V}$.

\subsubsection{Unified Auto-Regressive Next-Token-Prediction LMMs}

Developing the native long-multimodal thought process on top of unified auto-regressive next-token-prediction LMMs is a natural approach, since the model can intuitively generate the thought process token-by-token. However, the main technical challenge in implementing the native long-multimodal thought process on unified LMMs stems primarily from the developmental stage of the unified LMMs field itself. Such implementation requires models to inherently support interleaved multimodal generation--a frontier that remains less developed. Many current open-source models claiming to be ``unified vision understanding and generation LMMs'' typically lack native interleaved multimodal generation support, instead supporting only text or image outputs in isolation. Recent successes in closed-source models like GPT-4o and Gemini have demonstrated the tremendous potential of interleaved multimodal generation in real-world applications.

In this paper, we employ Anole \citep{chern2024anole, team2024chameleon} as our base model for generating native long-multimodal thought process for test-time scaling on LMMs. Anole is a unified auto-regressive next-token-prediction LMMs that are trained to directly predict the next multimodal (text or image) token. Anole presents several compelling advantages over alternative LMMs. First, Anole is pre-trained and post-trained directly on interleaved text-image tokens, enabling inherent capabilities for interleaved generation of multimodal tokens. Second, Anole's relatively efficient image representation encodes each image with 1024 tokens, making test-time scaling with the native long-multimodal thought process viable within reasonable inference compute budget. Third, Anole's modeling strategy closely resembles the state-of-the-art LLMs, making it capable to leverage existing training and inference infrastructure of LLMs.

\subsubsection{Methodology for Data Curation}
We deliberately curate our set of supervised fine-tuning (SFT) dataset, which consists of diverse vision (image) generation prompts to ensure high-quality alignment \citep{team2024chameleon}. To elicit LMMs to perform the native long-multimodal thought process, we carefully design and construct the solution multimodal reasoning chain to elicit LMMs' capabilities to spontaneously (1) critique their own generated visual steps and (2) generate intermediate visual subgoals. For more details, please refer to Section~\ref{sec:experiments}.

\section{Experiments}
\label{sec:experiments}

To instantiate \textit{Thinking with Generated Images} on vision generation tasks, we meticulously design two types of native long multimodal thought processes: \textit{vision generation with intermediate visual subgoals} and \textit{vision generation with self-critique}. We explore how these two processes can help LMMs to better conduct vision (image) generation via generative visual thinking.

\begin{figure}[ht]
    \centering
    \includegraphics[width=0.8\textwidth, height=8.3cm]{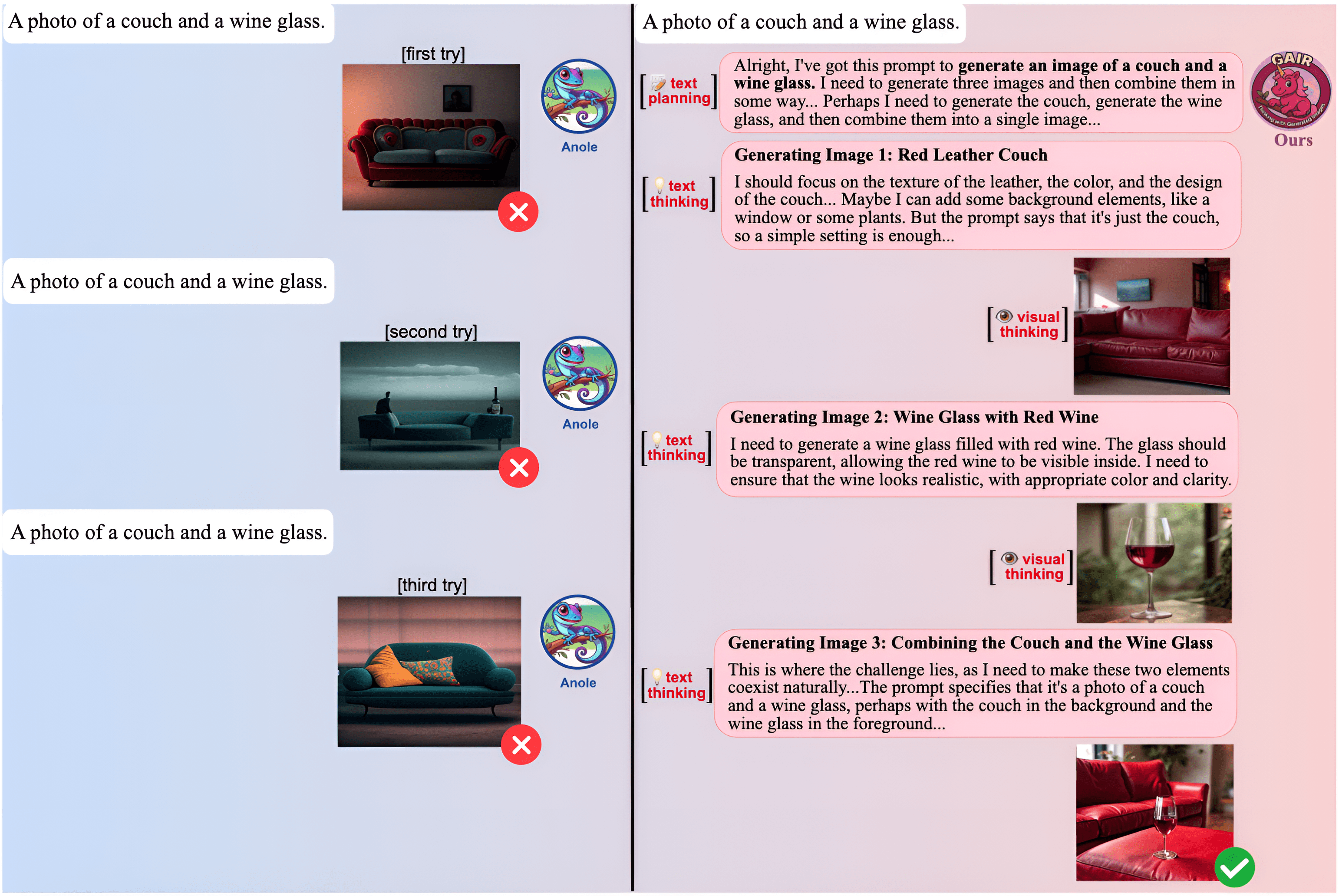}
    \caption{Demonstration of our long-multimodal thought process on GenEval.}
    \label{fig:geneval_example}
\end{figure}

\subsection{Vision Generation with Intermediate Visual Subgoals}
Through the \textit{vision generation with intermediate visual subgoals} process, LMMs are able to plan and achieve visual subgoals as intermediate steps toward the final solution, as presented in equation~\ref{eq:equation1}. Specifically, in the image generation task, when presented with a complex prompt containing multiple objects, the model generates each object individually and then integrates them into the final output image to fulfill the requirements of the multi-object prompt. An example shown in Fig.~\ref{fig:geneval_example} for demonstration.

\begin{equation}
\label{eq:equation1}
\text{[Text Planning]} \rightarrow \underbrace{\text{[Visual SubGoal $_i$]} \rightarrow \text{[Text Planning and Reflection]}}_{\text{iteratively}} \rightarrow \text{[Final Visual Output]}
\end{equation}

\subsection{Vision Generation with Self-Critique}
\begin{figure}[htb]
    \centering
    \includegraphics[width=0.75\textwidth, height=7.9cm]{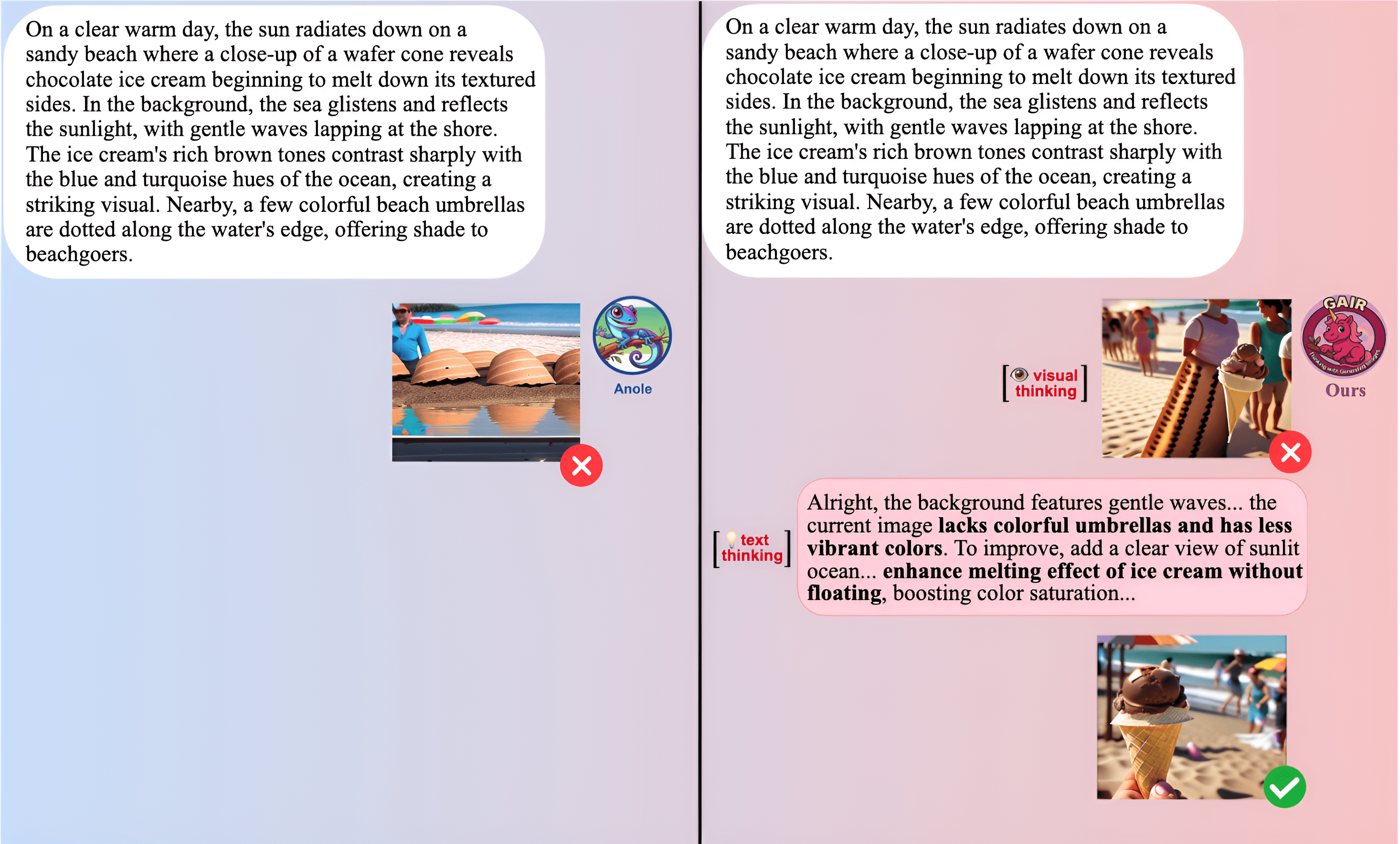}
    \caption{Demonstration of our long-multimodal thought process on DPG-Bench.}
    \label{fig:dpgbench_example}
\end{figure}

Through the \textit{vision generation with self-critique} process, LMMs are able to set a visual hypothesis, then perform critiquing and reflection the hypothesis, and finally make refinements toward better solution, as presented in equation~\ref{eq:equation2}. Specifically, in the image generation task, after generating an image based on a prompt, the model spontaneously reflects on the image to assess whether it meets the prompt's requirements. It then attempts to improve the image further, generating a higher-quality version that better aligns with the prompt by incorporating both the previously generated image and its own reflections. An example shown in Fig.~\ref{fig:dpgbench_example} for demonstration.

\begin{equation}
\label{eq:equation2}
\text{[Initial Visual Hypothesis]} \rightarrow  \text{[Text Planning and Reflection]} \rightarrow \text{[Final Visual Output]}
\end{equation}

\subsection{Dataset}
\label{subsec:dataset}
We carefully designed our synthetic SFT data–construction workflow that produces high-quality training samples to enable trained models to generate two types of long-multimodal thought processes. Specifically, we developed a pipeline that constructs synthetic examples illustrating long-multimodal reasoning. Our pipeline can generate the two types of long multimodal-thought process mentioned above. After fine-tuning on this data, the LMMs become capable of producing end-to-end long-multimodal chains of thought.

\begin{figure}[H]
    \centering
    \includegraphics[width=0.99\textwidth, height=3.6cm]{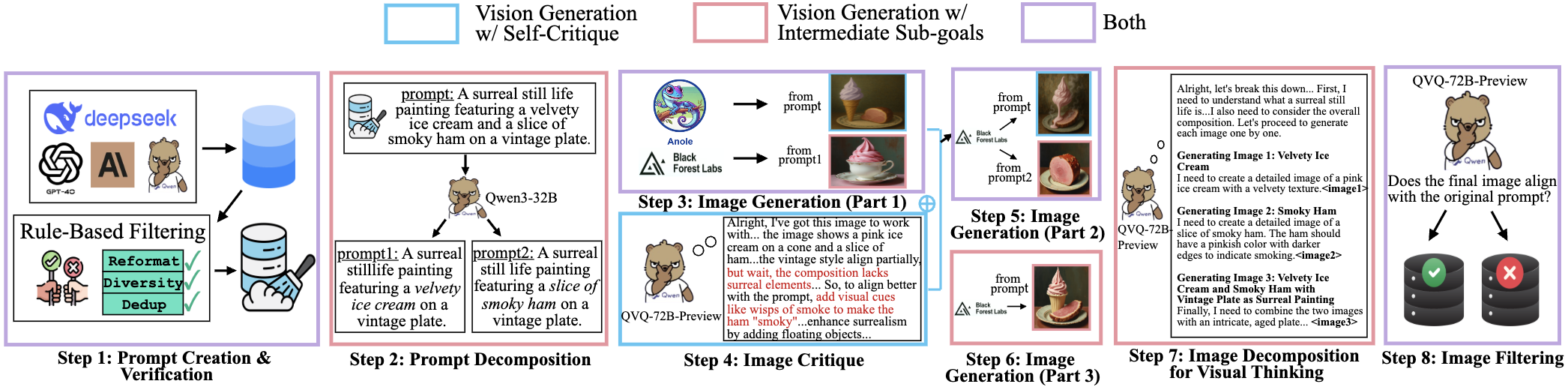}
    \caption{Our data collection pipeline for \textit{Thinking with Generated Images}.}
    \label{fig:data_pipeline}
\end{figure}

Since there are no single off-the-shelf LMM that supports dynamic test-time scaling, traditional distillation techniques are not applicable. Our full dataset-curation pipeline is illustrated in Fig.~\ref{fig:data_pipeline}.

We follow the below rules of thumb when collecting our synthetic data:

\begin{itemize}
\item \textbf{High-quality image-generation prompts:}
To ensure diverse, high-quality outputs, we leverage several state-of-the-art models (including Deepseek-V3 \citep{deepseekv3}, GPT-4o \citep{hurst2024gpt}, Claude3.7-Sonnet \citep{claude37sonnet}, and Qwen2.5-72B-Instruct \citep{qwen2.5}) to brainstorm complex prompts and then apply rule-based filtering to guarantee both variety and proper formatting.

\item \textbf{High-quality reflection-reasoning chains:}
We use QVQ-72B-Preview \citep{qwen2024qvq} to construct the core components of critique and reflection, conditioned on images.

\item \textbf{High-quality intermediate visual thoughts:}
We employ Flux1-dev \citep{flux2024} (for vision generation with intermediate subgoals) or Anole-7b \citep{chern2024anole} (for vision generation with self-critique) to generate initial visual hypotheses from the prompts. We then refine or update these with Flux1-Redux \citep{flux2024} (a variant of Flux1-dev that accepts both image and text inputs).
\end{itemize}

\subsubsection{High-Quality Image-Generation Prompts}
We curate diverse, complex prompts that go beyond simple object depiction to include variations in color, orientation, and shape with multiple elements. We deduplicate prompts \footnote{Prompt deduplication is crucial when leveraging language models for prompt generation, as these models sample from probability distributions that inherently lead to clustering around common outputs. When repeatedly sampling from these models, we observed occasional cases of repetition.} for data efficiency and reformat them into a unified structure optimized for our \textit{``thinking with generated images''} paradigm. For vision generation with intermediate subgoals, we use Qwen3-32B \citep{qwen2.5} to break down complex visual prompts into simpler components that can be processed independently.

\subsubsection{High-Quality Reflection-Reasoning Chains}
The image critique step employs QVQ-72B-Preview \citep{qwen2024qvq} for its strong long-CoT reasoning capabilities. For vision generation with self-critique, it analyzes each prompt-image pair, assessing accuracy, identifying discrepancies, and suggesting improvements. For vision generation with self-critique, it takes a holistic view of the entire image sequence, outputting structured reasoning chains that tries to reconstruct how a model could arrive at the final image through iterative decomposition. This process enables training our model to ``think visually''  by using images as a medium for solving tasks through reasoning.

\subsubsection{High-Quality Intermediate Visual Thoughts}
Our three-part image generation process creates visual representations of prompts. Initially, we use Anole-7b \citep{chern2024anole} (for vision generation with self-critique) or Flux1-dev \citep{flux2024} (for vision generation with intermediate subgoals) to generate first-round images. In the second stage, Flux1-Redux \cite{flux2024} refines images using a multimodal approach that incorporates the original prompt, first-stage image, and critique feedback to refine the image for the self-critique thought process. For the intermediate subgoals thought process, we instead use Flux1-Redux \cite{flux2024} conditioned on the first-round images (but based on the second component from the original prompt) to generate the second-round images. The third part of the image generation process continues by using Flux1-Redux \cite{flux2024} as the model (conditioned on both first and second round images) to generate a final image. Finally, QVQ-72B-Preview \cite{qwen2024qvq} performs quality control, filtering out examples where final generated images significantly deviate from the image-generation prompts.

\subsection{Training}

\paragraph{Loss Function Design}
There remains considerable room for optimization when training unified LMMs for visual generation tasks using only cross-entropy loss. To address image integrity issues arising from raster-based prediction, we introduce a reconstruction loss at the visual feature level. Specifically, we project the hidden states of the generated images back into the visual feature space and compute the mean squared error (MSE) loss against the corresponding features of the ground-truth image. This encourages the model to produce outputs with greater visual coherence and structural integrity.

Our ablation studies demonstrate that the model achieves optimal performance on the GenEval benchmark \citep{ghosh2023geneval} when the reconstruction loss is combined with the cross-entropy loss, using a weight of 1 for the reconstruction term. Thus, in subsequent training, we adopt a composite loss function that combines cross-entropy with the reconstruction loss (weighted at 1) to enhance the visual quality of the generated images. Further details on the auxiliary loss design and the effectiveness of this approach can be found in the Appendix.

\paragraph{Training Stages} The training is conducted in two stages. In the first stage, we performed continued training on Anole-7b with the JourneyDB dataset \citep{sun2023journeydb} to enhance the model's fundamental vision generation capabilities. In the second stage of training, we train the model with the synthetic dataset constructed as described in section~\ref{subsec:dataset}. We fine-tuned two models: TwGI-Anole-7b-Obj., using the vision generation with intermediate subgoals dataset, and TwGI-Anole-7b-Crit., using the vision generation with self-critique dataset. These two trained models are capable of generating visual intermediate subgoals and self-critiquing visual hypothesis, respectively.

\subsection{Inference}
Unlike standard VLMs or LLMs that don't need to think much about inference issues, for vision generation tasks on unified LMMs, classifier-free guidance \citep{ho2022classifier} will significantly improve model performance on vision generation. Specifically, on top of the full conditions, unconditions, and the image conditions as mentioned in \citep{team2024chameleon}, we add extra ``original prompt conditions'' and ``negative conditions'' to facilitate the intermediate visual steps to be more faithful without being disturbed too much by the generated long-text thought. By balancing between these conditions, we can enable the model to benefit from the long-text thought, without being distracted too much by potential noise inside the long thoughts.

\subsection{Main Results}
We benchmark the performance of TwGI-Anole-7b-Obj. and TwGI-Anole-7b-Crit. on GenEval \citep{ghosh2023geneval} and DPGBench \citep{hu2024ella}. We benchmark TwGI-Anole-7b-Obj. against the base model Anole-7b (which is also continued pretrained on the JourneyDB dataset), exploring whether generating visual subgoals can be beneficial for vision generation on complex prompts (i.e., in GenEval and DPGBench, this refers to image generation prompts that require generating at least two objects). Note that since some categories of GenEval require generating only single object, we omit the comparison on these categories. Then, we benchmark whether the TwGI-Anole-7b-Crit. model can correct its initial visual hypothesis (i.e., TwGI-Anole-7b-Crit. (visual hypo.) in Tab.~\ref{tab:geneval} and Tab.~\ref{tab:dpg}), and generate a better image generation result (i.e., TwGI-Anole-7b-Crit. (final) in Tab.~\ref{tab:geneval} and Tab.~\ref{tab:dpg}). 

\begin{table}[h]
    \centering
    \resizebox{\linewidth}{!}{
    \begin{tabular}{@{}l *{7}{c}}
    \toprule
    \multirow{3}{*}{\centering\textbf{Model}} & \multicolumn{7}{c}{\textbf{GenEval}}
    \\
    \cmidrule(lr){2-8}
     & \textbf{Single Obj.} & \textbf{Two Obj.} & \textbf{Counting} & \textbf{Colors} & \textbf{Position} & \textbf{Color Attri.} & \textbf{Overall$\uparrow$} \\
    \midrule
    LlamaGen \citep{sun2024autoregressive} & 0.71 & 0.34 & 0.21 & 0.58 & 0.07 & 0.04 & 0.32 \\
LDM \citep{rombach2022high} & 0.92 & 0.29 & 0.23 & 0.70 & 0.02 & 0.05 & 0.37 \\
Chameleon-7b \cite{team2024chameleon} & --  & --  & --  & -- & -- & -- & 0.39 \\
SDv1.5 \citep{rombach2022high} & 0.97 & 0.38 & 0.35 & 0.76 & 0.04 & 0.06 & 0.43 \\
    \midrule
    \textbf{Intermediate Visual SubGoals} &   &   &   &  &  &  & \\ 
    Anole-7b \citep{chern2024anole} & --  & 0.38  & --  & -- & 0.04 & 0.04 & -- \\
    \textbf{TwGI-Anole-7b-Obj.} & --  & \textbf{0.57}  & --  & -- & \textbf{0.11} & \textbf{0.16} & -- \\
    \midrule
    \textbf{Self-Critique on Visual Thoughts} &   &   &   &  &  &  & \\ 
    TwGI-Anole-7b-Crit. (visual hypo.) & 0.98 & 0.51 & 0.24 & 0.84 & 0.06 & 0.08 & 0.45 \\
    \textbf{TwGI-Anole-7b-Crit. (final)} & 0.98 & 0.59 & 0.23 & 0.87 & 0.12 & 0.10 & \textbf{0.48} \\
    \bottomrule
    \vspace{0.1mm}
    \end{tabular}
    }
    \caption{Performance on GenEval.}
    \label{tab:geneval}
\end{table}

\begin{table}
\centering
\footnotesize
\begin{tabular}{lccccccc}
\toprule
\textbf{Model} & \textbf{Global} & \textbf{Entity} & \textbf{Attribute} & \textbf{Relation} & \textbf{Other} & \textbf{Overall$\uparrow$} \\
\midrule
\textbf{Intermediate Visual SubGoals} &   &   &   &  &  &  \\ 
Anole-7b \citep{chern2024anole} & 74.16 & 70.31 & 69.42 & 80.12 & 43.60 & 58.32 \\
\textbf{TwGI-Anole-7b-Obj.} & 74.47 & 78.06 & 76.17 & 83.21 & 42.00 & \textbf{68.44} \\
\midrule
 \textbf{Self-Critique on Visual Thoughts} &   &   &   &  &  &  \\ 
 TwGI-Anole-7b-Crit. (visual hypo.) & 77.81 & 71.45 & 69.24 & 78.34 & 58.00 & 62.83 \\
 \textbf{TwGI-Anole-7b-Crit. (final)} & 76.90 & 75.70 & 73.88 & 81.86 & 55.60 & \textbf{67.14} \\
\bottomrule
\vspace{0.1mm}
\end{tabular}
\caption{Performance on DPG-Bench.}
\label{tab:dpg}
\end{table}

\subsection{Analysis}

\paragraph{Generating intermediate visual thoughts is beneficial for vision generation}
We show in Tab.~\ref{tab:geneval} and Tab.~\ref{tab:dpg} that TwGI-Anole-7b-Obj. consistently outperforms the baseline Anole-7b model across both GenEval and DPGBench. On GenEval, Anole-7b-Obj. achieves a significant boost in the "Two Obj." category (0.57 vs. 0.38), indicating its improved ability to handle complex prompts involving multiple entities. It also shows notable improvements in position and color attribute alignment, suggesting a stronger capacity for precise spatial and visual compositional reasoning. Similarly, on DPGBench, TwGI-Anole-7b-Obj. yields substantial gains in the "Entity", "Attribute", and "Relation" categories, reflecting its enhanced understanding of fine-grained visual semantics. These improvements validate our hypothesis that breaking down visual tasks into intermediate subgoals enables LMMs to reason more systematically and generate higher-quality outputs. 

\paragraph{The native long-multimodal thought process enables models to correct and refine their own visual hypothesis}
The results on \textit{self-critique on visual thoughts}, as shown in Tab.~\ref{tab:geneval} and Tab.~\ref{tab:dpg}, demonstrate the effectiveness of enabling models to reflect on and revise their own visual outputs. The TwGI-Anole-7b-Crit. model achieves a notable increase in performance after the self-critique step, improving the overall GenEval score from 0.45 to 0.48, and the DPG-Bench score from 62.83 to 67.14. This indicates that the ability to introspectively analyze generated images—through textual reasoning chains grounded in visual feedback—allows the model to identify mismatches, hallucinations, or missing elements, and subsequently correct them. The effectiveness of this visual feedback loop reflects a form of inter-modality synergy, where visual and textual modalities iteratively guide each other. We further include qualitative examples in the Appendix to illustrate how critique-driven revisions result in visually and semantically improved generations.

\section{Conclusion}
In this paper, we introduce the concept of \textit{Thinking with Generated Images}, and implement it with the \textit{native long-multimodal thought process} on autoregressive next-token-prediction LMMs. While we focus on text and vision modalities, our core idea can be extended to diverse modalities. As more capable LMMs continue to emerge, particularly in the open-source domain, with the \textit{Thinking with Generated Images} paradigm, we anticipate future AI models will explore protein structures or revise building designs as naturally as they write a poem.

\section{Limitations and Future Directions}\label{sec:limitandfuture}

\subsection{Limitations}
In this paper, we demonstrate \textit{Thinking with Generated Images} on Anole-7b \citep{chern2024anole}. As mentioned in the introduction, the developmental stage (especially in the open-source realm) of unified LMMs is still evolving. With stronger unified LMMs emerging in the future, we anticipate more powerful or even emergent capabilities via the \textit{Thinking with Generated Images} paradigm. Also, while in this paper we demonstrate our approach on autoregressive next-token-prediction LMMs, we anticipate our core idea can be applied to diffusion-based LMMs or mixed autoregressive/diffusion LMMs. We leave the exploration on different architectures to future work.

\subsection{Future Directions}

\paragraph{Better Benchmarking on Thinking with Generated Images}
Current vision generation benchmarks for unified LMMs focus on standard image generation tasks. In this paper, we also use standard image generation benchmarks \citep{ghosh2023geneval,hu2024ella}. However, with the increased inherent capabilities of LMMs and emergent capabilities coming to light, real-world tasks such as those illustrated in Fig.~\ref{fig:framework} and Fig.~\ref{fig:see_vs_think} will become increasingly feasible for LMMs via \textit{Thinking with Generated Images}. We need more realistic benchmarks to evaluate these models.

\paragraph{Test-Time and Post-Training Scaling on Unified LMMs}
Our work represents the first step for test-time scaling on unified LMMs. As stronger unified LMMs emerge, we believe that test-time scaling and post-training scaling will become more viable, effective, and worth exploring.

\paragraph{Efficient Vision Representation for LMMs} Efficient vision representation will be essential for achieving scalable test-time and post-training scaling in the vision modality. Recent research demonstrates that images can be effectively represented with as few as 32 or even 16 tokens/patches \citep{yu2025image, zhou2024transfusion}. We believe this line of work is a promising direction for future research.

\section{Acknowledgment}
We express our sincere gratitude to Mingxuan Wang for his valuable support on this work.
This project is supported by SJTU SEIEE - ByteDance Large Language Model Joint Laboratory and SII.

\newpage

\bibliographystyle{acl_natbib}
\bibliography{main}

\newpage

\appendix

\section{Experiment Details}

\subsection{Training Loss}
This section details our loss function design, with particular emphasis on the visual modality in multi-modal training samples. Given a batch of sequences that may contain multiple images interleaved with text, our model processes each sequence through a backbone transformer to obtain hidden representations:
$$\mathbf{h} = f_{\text{trans}}(\mathbf{x}; \theta)$$
where $\mathbf{x}$ represents the tokenized input sequence and $\mathbf{h} \in \mathbb{R}^{B \times L \times D}$ denotes the hidden states for batch size $B$, sequence length $L$, and hidden dimension $D$.

The model employs a unified multimodal vocabulary space where image and text tokens share the same embedding space. Specifically, vocabulary indices 4 through 8195 (8192 tokens total) correspond to visual codebook entries from the VQ-VAE, while the remaining indices represent text tokens. A multimodal language modeling head produces logits over this joint vocabulary:
$$\mathbf{z} = f_{\text{mmlm}}(\mathbf{h}; \theta_{\text{mmlm}})$$
where $\mathbf{z} \in \mathbb{R}^{B \times L \times V}$ and $V$ is the total vocabulary size encompassing both modalities.

Our training objective combines two loss components:

\textbf{Autoregressive Multimodal Loss}: We compute the standard cross-entropy loss over all tokens (both text and image) in the sequence:
$$\mathcal{L}_{\text{mm}} = -\frac{1}{|\mathcal{T}|} \sum_{(b,i) \in \mathcal{T}} \log p(x_{b,i+1} | \mathbf{x}_{b,1:i})$$
where $(b,i) \in \mathcal{T}$ denotes valid token positions with $b \in \{1, \ldots, B\}$ being the batch index and $i \in \{1, \ldots, L_b-1\}$ being the token position within sequence $b$. The probability $p(x_{b,i+1} | \mathbf{x}_{b,1:i})$ is computed by applying softmax to the logits $\mathbf{z}_{b,i} \in \mathbb{R}^V$ at position $i$, which are generated by the transformer conditioned on all previous tokens $\mathbf{x}_{b,1:i}$. The set $\mathcal{T}$ excludes padded positions and positions where ground-truth labels are masked. This loss naturally handles both text token prediction and image token generation within the unified vocabulary space.

\textbf{Visual Reconstruction Loss}: For image tokens specifically, we extract visual features and compute a reconstruction objective. Each image in the input is demarcated by special tokens $\langle \text{BOI} \rangle$ (beginning-of-image, token ID 8197) and $\langle \text{EOI} \rangle$ (end-of-image, token ID 8196), with exactly 1024 tokens between them. For each image $j$ in the batch:
\begin{enumerate}
\item Extract the corresponding hidden states: $\mathbf{h}^{(j)}_{\text{img}} \in \mathbb{R}^{1024 \times D}$
\item Project to codebook space: $\mathbf{p}^{(j)} = f_{\text{proj}}(\mathbf{h}^{(j)}_{\text{img}}; \theta_{\text{proj}})$, where $f_{\text{proj}}$ is an additional trainable linear projection layer that maps from hidden dimension $D$ to codebook dimension $D'$, i.e., $f_{\text{proj}}: \mathbb{R}^D \rightarrow \mathbb{R}^{D'}$
\item Retrieve original codebook features: $\mathbf{c}^{(j)} = f_{\text{vq}}(\text{image}_j; \theta_{\text{vq}})$, where $\mathbf{c}^{(j)} \in \mathbb{R}^{1024 \times D'}$ represents the quantized codebook features for the $j$-th image
\end{enumerate}

The visual reconstruction loss is computed as the mean squared error (MSE) between the projected hidden states and the original codebook features:
$$\mathcal{L}_{\text{rec}} = \frac{1}{N_{\text{img}}} \sum_{j=1}^{N_{\text{img}}} \frac{1}{1024 \cdot D'} \|\mathbf{p}^{(j)} - \mathbf{c}^{(j)}\|_2^2$$
where $N_{\text{img}}$ is the total number of images across all sequences in the batch. Note that if a batch contains no images, $\mathcal{L}_{\text{rec}} = 0$.

The total loss is:
$$\mathcal{L}_{\text{total}} = \mathcal{L}_{\text{mm}} + \lambda \mathcal{L}_{\text{rec}}$$

Previous works have made initial attempts to train autoregressive next-token-prediction LMMs beyond cross-entropy loss \citep{li2025imagine, hu2025fine}. To rigorously benchmark the effectiveness of incorporating the visual reconstruction loss, we fine-tuned Anole-7b on 50,000 images from JourneyDB and evaluated on the GenEval benchmark. We investigated different values of $\lambda$ and compared against the token discrepancy loss $\mathcal{L}_{\text{td}}$ proposed by \citep{li2025imagine}. Note that we did not apply classifier-free guidance to ensure fair and simple comparison. Results in Tab.~\ref{tab:loss_combined} demonstrate that $\lambda = 1$ yields optimal performance, improving GenEval scores by approximately 3 points. Neither $\mathcal{L}_{\text{td}}$ alone nor the combination $\mathcal{L}_{\text{mm}} + \lambda \mathcal{L}_{\text{rec}} + \mu \mathcal{L}_{\text{td}}$ surpassed our proposed objective. Thus, we adopt $\mathcal{L}_{\text{total}} = \mathcal{L}_{\text{mm}} + \mathcal{L}_{\text{rec}}$ with $\lambda = 1$ for all experiments in this paper.

\begin{table}[h!]
    \centering
    \resizebox{\linewidth}{!}{
    \begin{tabular}{@{}l *{7}{c}}
    \toprule
    \multirow{3}{*}{\centering\textbf{Model}} & \multicolumn{7}{c}{\textbf{GenEval}}
    \\
    \cmidrule(lr){2-8}
     & \textbf{Single Obj.} & \textbf{Two Obj.} & \textbf{Counting} & \textbf{Colors} & \textbf{Position} & \textbf{Color Attri.} & \textbf{Overall$\uparrow$} \\
    \midrule
    \textbf{Baseline} &   &   &   &  &  &  & \\ 
    $L_{mm}$ & 0.84 & 0.20 & 0.11 & 0.59 & 0.02 & 0.04 & 0.30 \\
    \midrule
    \textbf{Reconstruction Loss Variants} &   &   &   &  &  &  & \\ 
    $L_{mm}+0.5\cdot L_{rec}$ & 0.86 & 0.21 & 0.15 & 0.63 & 0.04 & 0.06 & 0.32 \\
    $L_{mm}+L_{rec}$ & 0.86 & 0.26 & 0.12 & 0.61 & 0.04 & 0.08 & \textbf{0.33} \\
    $L_{mm}+5\cdot L_{rec}$ & 0.86 & 0.22 & 0.12 & 0.58 & 0.03 & 0.04 & 0.31 \\
    \midrule
    \textbf{Token Discrepancy Loss Variants} &   &   &   &  &  &  & \\ 
    $L_{mm}+0.5\cdot L_{td}$ & 0.84 & 0.24 & 0.11 & 0.58 & 0.02 & 0.05 & 0.31 \\
    $L_{mm}+L_{td}$ & 0.84 & 0.21 & 0.13 & 0.60 & 0.03 & 0.05 & 0.31 \\
    $L_{mm}+5\cdot L_{td}$ & 0.80 & 0.18 & 0.10 & 0.56 & 0.02 & 0.04 & 0.28 \\
    \midrule
    \textbf{Combined Loss} &   &   &   &  &  &  & \\ 
    $L_{mm}+L_{td}+L_{rec}$ & 0.88 & 0.22 & 0.13 & 0.56 & 0.06 & 0.03 & 0.31 \\
    \bottomrule
    \end{tabular}
    }
    \vspace{4mm}
    \caption{Evaluation results on GenEval with different loss functions.}
    \label{tab:loss_combined}
\end{table}

\subsection{Training Stages Details}
Tab.~\ref{tab:training_details} shows the hyperparameters used in our training experiments. For stage-1 training, we train on approximately 4M text-image pairs. For stage-2 training, we use approximately 5k samples for intermediate visual subgoals and 40k samples for self-critique on visual thoughts, respectively.

\begin{table}[h!]
\centering
\resizebox{0.8\linewidth}{!}{
\begin{tabular}{@{}l *{4}{c}}
\toprule
\multirow{2}{*}{\textbf{Hyperparameters}} & \multicolumn{2}{c}{\textbf{Intermediate Visual Subgoals}} & \multicolumn{2}{c}{\textbf{Self-Critique on Visual Thoughts}} \\
\cmidrule(lr){2-3} \cmidrule(lr){4-5}
& \textbf{Stage 1} & \textbf{Stage 2} & \textbf{Stage 1} & \textbf{Stage 2} \\
\midrule
Learning rate & $1.0 \times 10^{-5}$ & $1.0 \times 10^{-5}$ & $1.0 \times 10^{-5}$ & $1.0 \times 10^{-5}$ \\
LR scheduler & Linear & Linear & Linear & Linear \\
Gradient clip & 1.0 & 1.0 & 1.0 & 1.0 \\
Optimizer & \multicolumn{2}{c}{AdamW ($\beta_1 = 0.9$, $\beta_2 = 0.999$)} & \multicolumn{2}{c}{AdamW ($\beta_1 = 0.9$, $\beta_2 = 0.999$)} \\
Training steps & 5K & 2K & 5K & 26K \\
Batch size & 1536 & 8 & 1536 & 8 \\
\bottomrule
\end{tabular}
}
\vspace{4mm}
    \caption{Hyperparameters for training \textit{vision generation with intermediate visual subgoals} and \textit{vision generation with self-critique}. Note that \textit{vision generation with intermediate visual subgoals} and \textit{vision generation with self-critique} share the same stage-1 model. The stage-1 model is then separately fine-tuned for generating different types of native long-multimodal thought process.}
    \label{tab:training_details}
\end{table}

\subsection{Inference Details}
Classifier-free guidance (CFG) is crucial for enhanced image generation quality. We apply the following CFG settings to ensure balanced conditioning, enabling the model to concurrently leverage text-based thoughts, previous visual hypothesis, visual subgoals, and original prompts. Specific configurations are shown in Tab.~\ref{tab:cfg_scale}.

\begin{table}[ht]
\centering
\resizebox{0.8\linewidth}{!}{
\begin{tabular}{@{}l *{3}{c}}
\toprule
\textbf{CFG scale} & \multicolumn{2}{c}{\textbf{Intermediate Visual Subgoals}} & \textbf{Self-Critique on Visual Thoughts} \\
\cmidrule(lr){2-3}
& \textbf{subgoals} & \textbf{final} & \\
\midrule
Full conditions & 5.0 & 2.0 & 1.5 \\
Image conditions & 0.0 & 1.2 & 0.8 \\
Negative conditions & 3.0 & 3.0 & 3.0 \\
Prompt conditions & 0.0 & 5.0 & 5.0 \\
\bottomrule
\end{tabular}
}
\vspace{4mm}
    \caption{CFG scale for \textit{vision generation with intermediate visual subgoals} and \textit{vision generation with self-critique}.}
    \label{tab:cfg_scale}
\end{table}

\newpage
\section{Case Study}

\subsection{Vision generation with intermediate visual subgoals}

\begin{figure}[ht]
    \centering
    \includegraphics[width=0.95\textwidth, height=11cm]{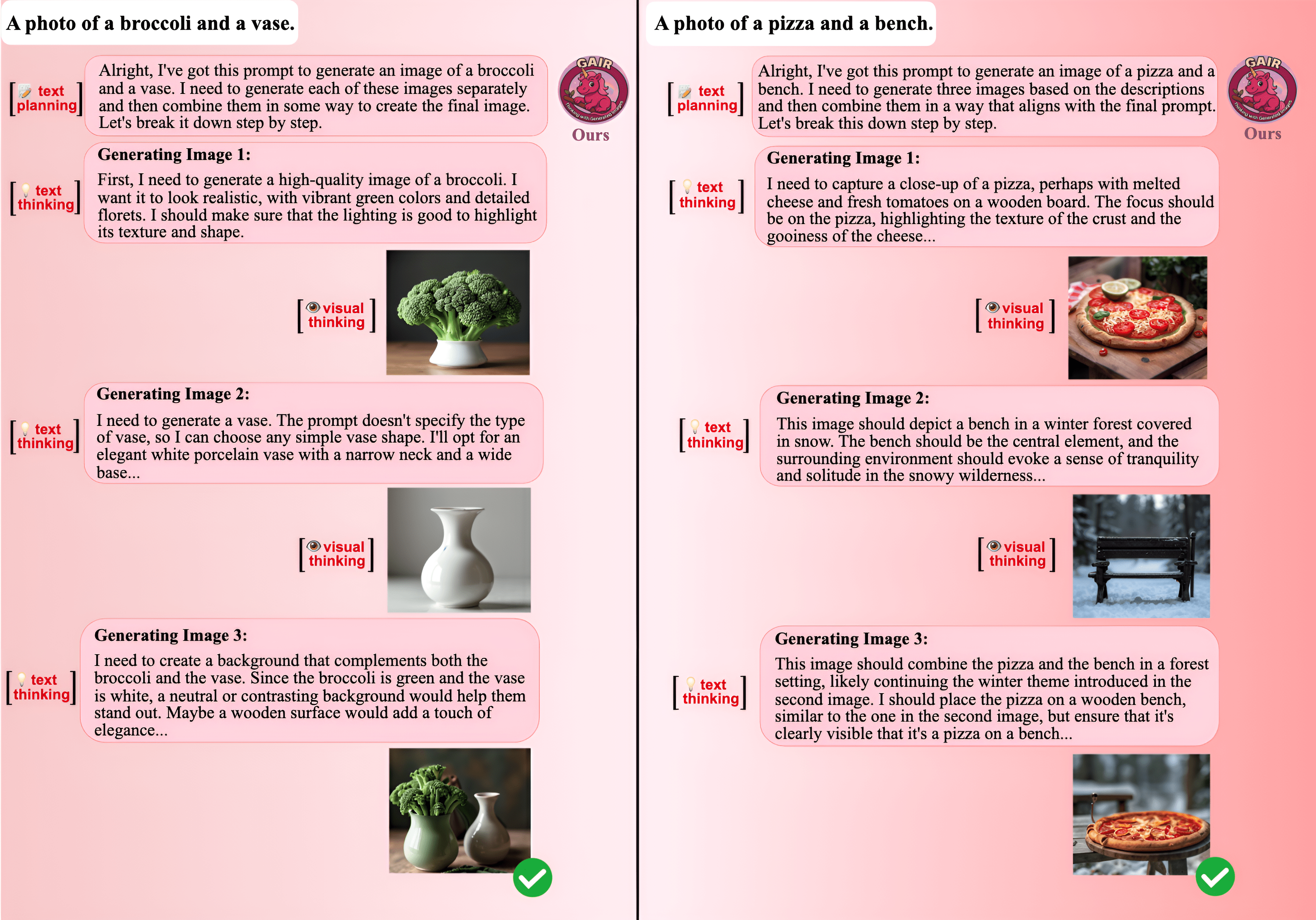}
    \caption{\textit{Vision generation with intermediate visual subgoals} on GenEval.}
    \label{fig:subgoal}
\end{figure}

Fig.~\ref{fig:subgoal} shows cases of vision generation with intermediate subgoals. The model demonstrates its ability to decompose complex visual generation tasks into manageable sub-components. When generating \textit{broccoli and vase}, 
the model first generates a high-quality image of a broccoli, focusing on realistic textures and vibrant green colors with detailed florets. It then generates a simple vase. In the final step, it combines these elements to create the complete image as requested by the prompt. Similarly, when tasked with generating both a \textit{pizza and a bench}, the model first creates each object separately, producing a detailed pizza with visible toppings, and then a wooden bench. The final image successfully integrates both elements, demonstrating the model's ability to handle multiple distinct objects in a single composition.
Generating intermediate visual subgoals allows the model to focus on completing each subgoal before producing the final output, resulting in a higher-quality solution.

\newpage

\subsection{Vision generation with self-critique}

\begin{figure}[ht]
    \centering
    \includegraphics[width=0.9\textwidth, height=11cm]{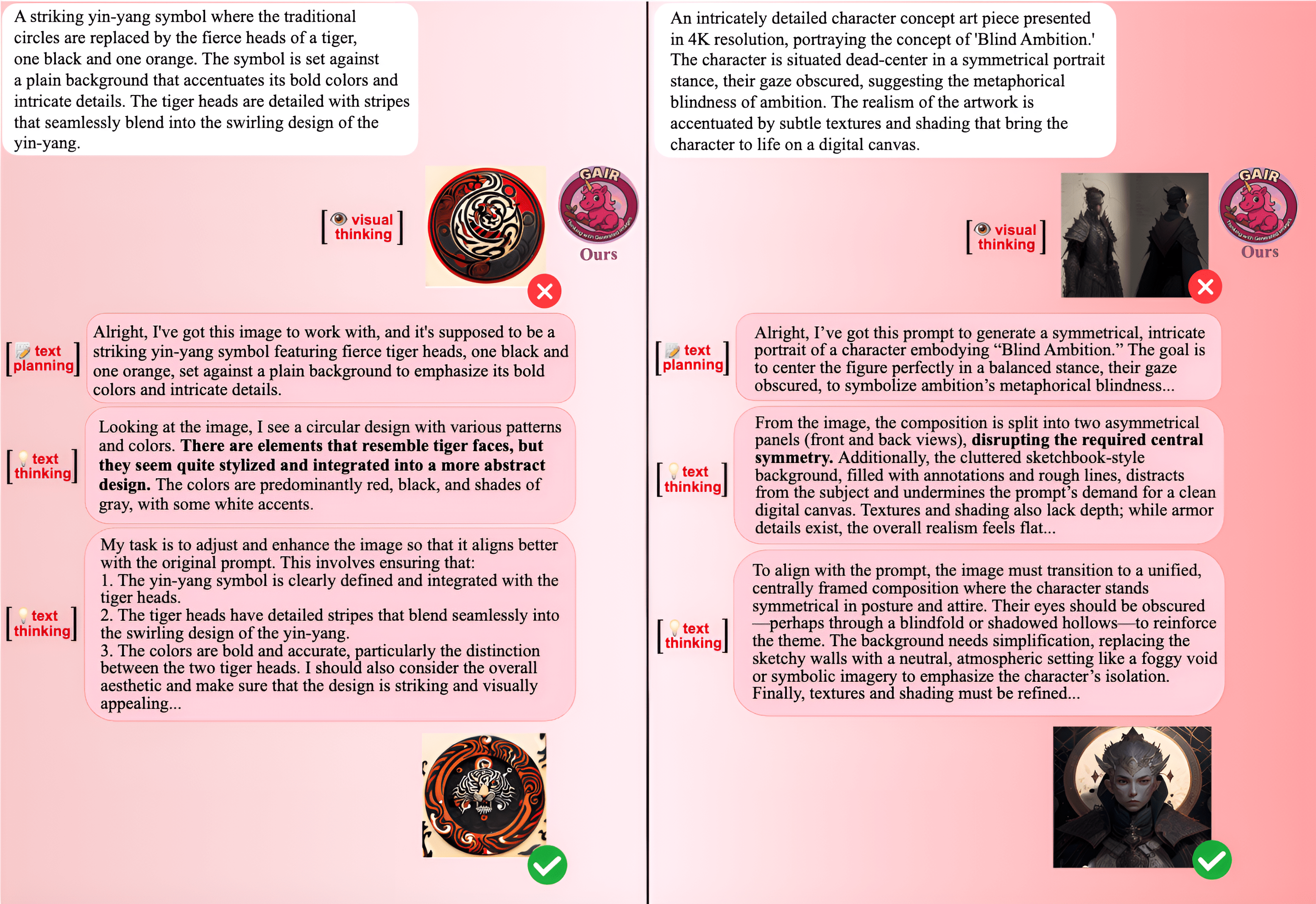}
    \caption{\textit{Vision generation with self-critique} on DPG-Bench.}
    \label{fig:critique}
\end{figure}

Fig.~\ref{fig:critique} shows cases of vision generation with self-critique. The model demonstrates its ability to analyze initial visual hypothesis and iteratively improve them through reflection. For generating a yin-yang symbol with a tiger head, the model initially generates a imperfect yin-yang symbol featuring a tiger. Through self-critique, it recognizes the need to emphasize the tiger faces to be more stylized and integrated. In the final version, the model successfully implements the corrections, with the tiger head now seamlessly blend into the yin-yang design. For generating a 4K-resolution portrait of “Blind Ambition,” the model first produces a split-view sketch with rough annotations, uneven shading, and a cluttered background that breaks the required central symmetry and leaves the character’s gaze only loosely obscured. Upon self-critique, it recognizes that the composition must be unified and centrally framed, the metaphorical blindness needs either a clear blindfold or shadowed hollows, and the background should shift from sketchy walls to a neutral, atmospheric void while textures demand deeper, more subtle rendering. The final image successfully centers a lone figure in a balanced stance, with warm steel armor carved by refined gradients, and a misty, minimalist backdrop that underscores the character’s solitary, driven ambition.

\newpage

\subsection{Failure Cases}

\begin{figure}[ht]
    \centering
    \includegraphics[width=0.9\textwidth, height=13cm]{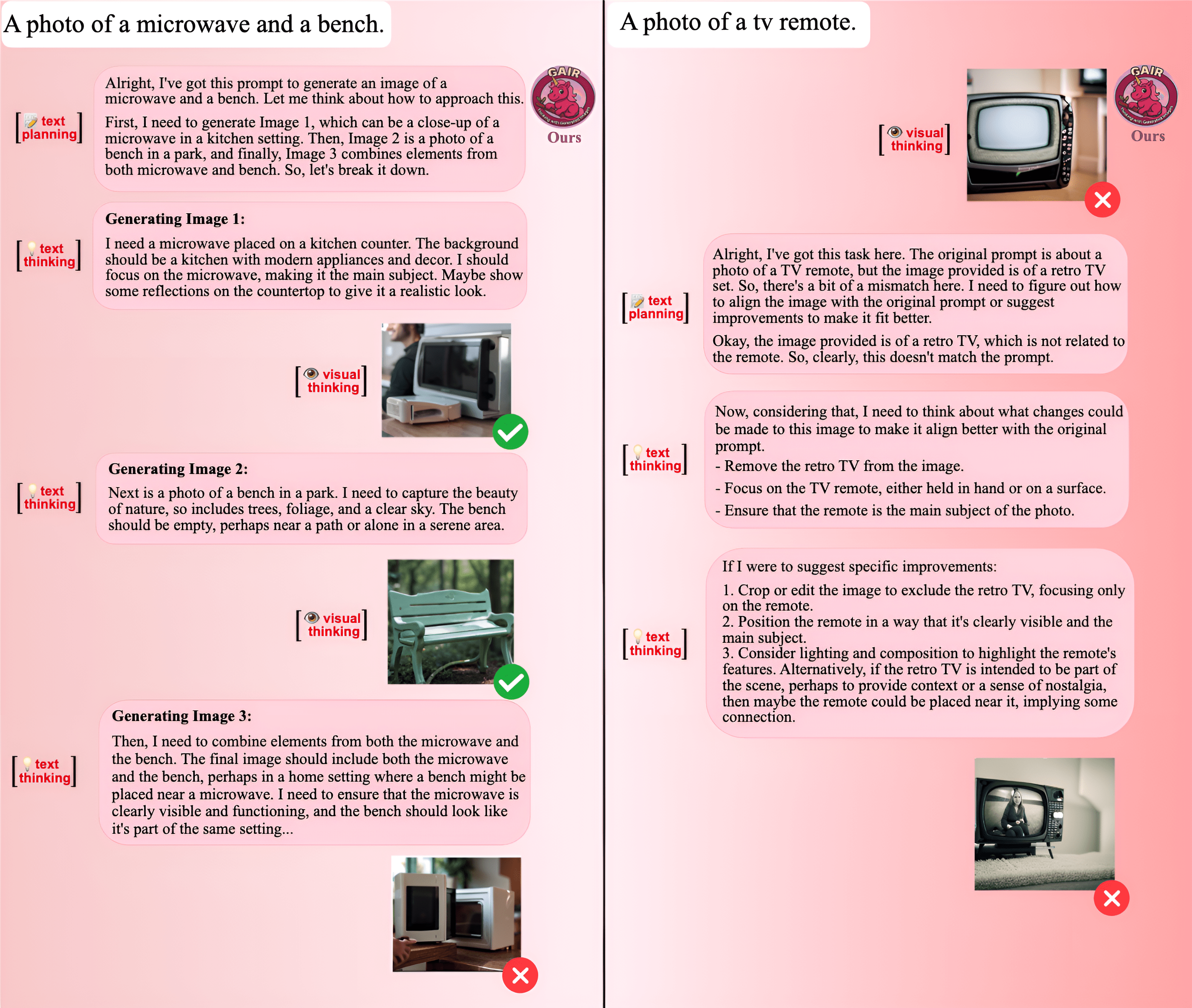}
    \caption{Example failure cases of using \textit{thinking with generated images}. Either the intermediate images generated are correct but fails to combine them together in the end, or the critique is correct but didn't generate the corresponding correct image.}
    \label{fig:failure}
\end{figure}

The left-half of Fig.~\ref{fig:failure} shows a failure case of vision generation with intermediate subgoals. While the model successfully generates individual components, it fails to combine them in the final output. When tasked with generating \textit{microwave and bench}, the model first produces a microwave placed on a kitchen counter, then generates a bench in a park setting. However, in the final step, the model fails to meaningfully combine both elements. This failure illustrates that while the model can correctly generate individual subgoals (the microwave and the bench separately), it sometimes struggles with the spatial reasoning required to integrate disparate objects that don't naturally belong together. We anticipate this issue to be mitigated with stronger base models or further post-training.

The right-half of Fig.~\ref{fig:failure} shows a failure case of vision generation with self-critique. When prompted to generate \textit{a photo of a tv remote}, the model initially produces a retro TV set instead, completely misinterpreting the request. Through self-critique, the model correctly identifies this error and provides detailed suggestions: remove the TV, focus on the remote either held in hand or on a surface, and ensure the remote is the main subject. Despite this accurate analysis and reasonable corrective suggestions, the second attempt still shows a TV set rather than a TV remote. This failure demonstrates a disconnect between the model's analytical capabilities and generative execution—while the self-critique mechanism can accurately identify problems and propose solutions, the model sometimes fails to implement its own corrections in subsequent generations. The model can reason about what's wrong but cannot translate this understanding into proper image generation. Similarly, we anticipate this issue to be mitigated with stronger base models or further post-training.

\end{document}